# Knowledge-Driven Vision-Language Model for Plexus Detection in Hirschsprung's Disease

Youssef Megahed[*]
Department of Systems and Computer Engineering
Carleton University
Ottawa, Ontario, Canada
youssefmegahed@cmail.carleton.ca

Atallah Madi[*]
Department of Systems and Computer Engineering
Carleton University
Ottawa, Ontario, Canada
atallahmadi@cmail.carleton.ca

Dina El Demellawy
Department of Pathology
Children's Hospital of Eastern Ontario (CHEO)
Ottawa, Ontario, Canada
deldemellawy@cheo.on.ca

Adrian D. C. Chan
Department of Systems and Computer Engineering
Carleton University
Ottawa, Ontario, Canada
Adrian.Chan@carleton.ca

## Abstract

Hirschsprung's disease is defined as the congenital absence of ganglion cells in some segment(s) of the colon. The muscle cannot make coordinated movements to propel stool in that section, most commonly leading to obstruction. The diagnosis and treatment for this disease require a clear identification of different region(s) of the myenteric plexus, where ganglion cells should be present, on the microscopic view of the tissue slide. While deep learning approaches, such as Convolutional Neural Networks, have performed very well in this task, they are often treated as black boxes, with minimal understanding gained from them, and may not conform to how a physician makes decisions. In this study, we propose a novel framework that integrates expert-derived textual concepts into a Contrastive Language-Image Pre-training-based vision-language model to guide plexus classification. Using prompts derived from expert sources (e.g., medical textbooks and papers) generated by large language models and reviewed by our team before being encoded with QuiltNet, our approach aligns clinically relevant semantic cues with visual features. Experimental results show that the proposed model demonstrated superior discriminative capability across different classification metrics as it outperformed CNN-based models, including VGG-19, ResNet-18, and ResNet-50; achieving an accuracy of 83.9%, a precision of 86.6%, and a specificity of 87.6%. These findings highlight the potential of multi-modal learning in histopathology and underscore the value of incorporating expert knowledge for more clinically relevant model outputs.

## 1 Introduction

Hirschsprung disease (HD), a congenital birth disorder, causes colonic nerve cell deformation and intestinal blockage [21]. HD affects 1 in 5000 newborns globally and may be deadly if untreated [1]. A lack of colon ganglion cells characterizes HD [21]. These ganglion cells are located in the myenteric plexus, particularly in the muscularis propria part of the colon (Fig. 1). Pull-through surgery, a common surgical treatment, removes the aganglionic colon and attaches the healthy part to the anus [10].

Pathologists visually examine colon histopathology images to identify healthy and aganglionic sections. More quantitative analyses, such as counting the ganglion cells and assessing their spatial distribution, may improve surgical success and patient outcomes by establishing more criteria for identifying healthy and aganglionic colon sections, but they would take time and increase healthcare costs for pathologists [10]. In addition, manual assessment is prone to inter- and intra-rate variability [2, 3] since different pathologists would label the cells as ganglion cells differently. This motivates the need for a computational tool that can assist or automate the quantification of ganglion cells to support a more consistent and efficient diagnostic process.

Efforts have been undertaken to automate the examination of histopathological whole slide images (WSIs) of the colon, whereby the identification of ganglion cells was categorized into a three-stage process [4–7, 10]: (1) Segmenting the muscularis; (2) inside the muscularis, delineating the plexus regions; and (3) within the plexus regions, identifying ganglion cells.

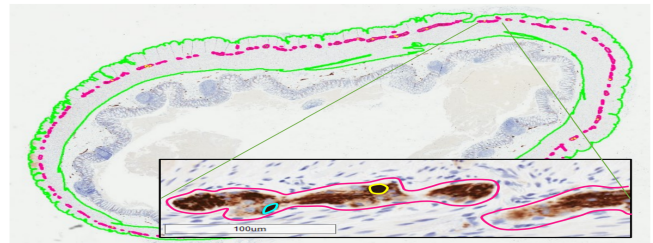

**Figure 1: Annotated histological section with the muscularis propria (green), myenteric plexus (pink), high-confidence ganglia (yellow), and low-confidence ganglia (cyan) [7].**

In prior works for stage (2), a shallow machine learning approach known as k-means clustering was used, and it achieved a Precision score of approximately 73.8%, Recall of 85.9%, and a Ganglia Inclusion Rate (GIR) of 99.2% [7]. GIR is the percentage of all ganglion cells that are found within the segmented plexus region. This is a valuable measure, as not all plexus segmentation errors are detrimental. If the segmented plexus still contains all the ganglion cells, the segmentation can be considered acceptable because, ultimately, we are interested in identifying the ganglion cells, which are found in these plexus regions.

---

[*]Both authors contributed equally to this research.

Recently, deep learning (DL) approaches have demonstrated a state-of-the-art performance in medical image analysis, including computational pathology [10]. Convolutional neural networks (CNNs) are one of the popular deep-learning approaches that are used for stage (1) with 81.9% Precision, and Recall of 96.2% [6, 7].

All these automated methods use semantic pixel-level segmentation. They have achieved promising results and advanced the field, but they only use visual features from histopathology images. This image-centric approach does not fully align with how pathologists typically classify Hirschsprung's disease. Clinical diagnosis requires visual identification of ganglion cells and contextual information like tissue structure, patient history, and spatial relationships between anatomical structures. This limitation can cause model misattribution, as in the famous husky dog and wolf classification case, where images were found to be discerned based on whether or not the background of the image contained snow [9]. In such a case, the system relies on spurious correlations instead of relevant features. This gap emphasizes the need for domain-specific knowledge and nuanced reasoning as human experts, where we use true disease indicators.

This study proposes a methodology that augments a Contrastive Language-Image Pre-training (CLIP)-based Vision-Language Model (VLM) with expert-derived concepts to guide the classification of plexus regions within the muscularis propria layer. In contrast to previous attempts, this approach frames the task as a classification problem; to ensure accurate identification of all ganglia cells within these plexus regions, aligning the model's reasoning for effective diagnosis of Hirschsprung's disease with a process similar to the human expert diagnosis rationale.

In this work, we introduce a knowledge-driven VLM that integrates expert-derived textual concepts with histopathological image features for the classification of plexus regions in Hirschsprung's disease. Our method integrates domain knowledge into the decision-making process by using LLMs to extract clinically validated descriptors and aligning them with visual representations from a CLIP-based backbone. This framework enhances interpretability and classification accuracy by integrating model reasoning with clinical diagnostic processes.

## 2 Methodology
### 2.1 Data Overview
The proposed study utilized a dataset that was employed in recent HD research [4, 7, 10]. The dataset consisted of 30 WSIs from 26 patients diagnosed with HD. Images were scanned using the digital scanner Aperio Scan Scope CS at 20× resolution (0.50 µm/pixel) by the Children's Hospital of Eastern Ontario (CHEO) [7]; to produce a digital version of the whole slide sample. The size of WSIs ranged from 17,983×17,602 to 65,940×34,997 pixels, with three colour channels. This dataset is from a closed source and is only accessible to the members found on the Carleton University-approved Research Ethics Boards (REB) list.

For each WSI, there is three manually annotated ground truth images that follows a hierarchical structure, where each can be found within the other. These ground truth images are (i) the muscularis propria, (ii) the myenteric plexus regions (i.e., an amount of the surrounding tissue around the plexus regions were included) shown in Fig. 2, and (iii) the ganglion cells (i.e., while some parts of the ganglion cell might be excluded, some areas surrounding the cell that can be included, as it can be challenging to define the cell's boundaries) [10]. In addition to the ground truth images, each WSI has a boundary outline for each of the three tissue structures with confidence level for each ganglion cell annotation which can be seen in Fig. 1 [7]. High confidence level reflect the strong indication of such annotation while lower level indicates uncertainty.

### 2.2 Data Preprocessing
There are preprocessing steps that were applied to the entire dataset to ensure consistency and to generate suitable inputs for model training and evaluation of the classification task.

First, colour normalization was performed using the Macenko method, a widely adopted stain normalization technique in computational pathology [8]. This eliminates the colour variation found in each WSI, which results in reducing inter-slide and inter-patient staining variability caused by the different histological staining process or scanning conditions.

Second, 20× WSIs were downsampled to 5×. This reduction balances the tradeoff between resolution and field-of-view. Only a small part of the plexus region is covered by a 224×224 tile at 20×, frequently lacking contextual information about the surrounding area. Nevertheless, at 5×, the same tile size covers a wider field-of-view, which enables the model to view the entire plexus region in a single patch with enough resolution for precise classification. Subsequently, overlapping tiles of 224×224 pixels were extracted from each WSI to create compatible inputs for the DL models to be tested. The tiles were extracted at a fixed stride of 112 pixels, corresponding to a 50% overlap between adjacent tiles to ensure contextual continuity and adequate coverage, especially at region boundaries.

Classification labels were derived from the manually annotated ground truth segmentation maps (Fig. 2). Each tile was labelled as "plexus" if at least one pixel overlapped with the manually annotated myenteric plexus regions; otherwise, it was labelled as "no plexus." This binary classification approach allows the model

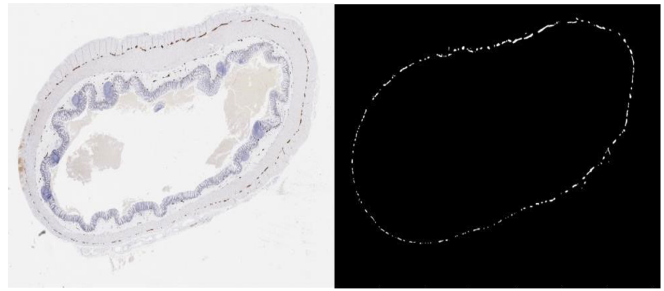

**Figure 2: A WSI sample with its corresponding ground truth mask for plexus regions.**

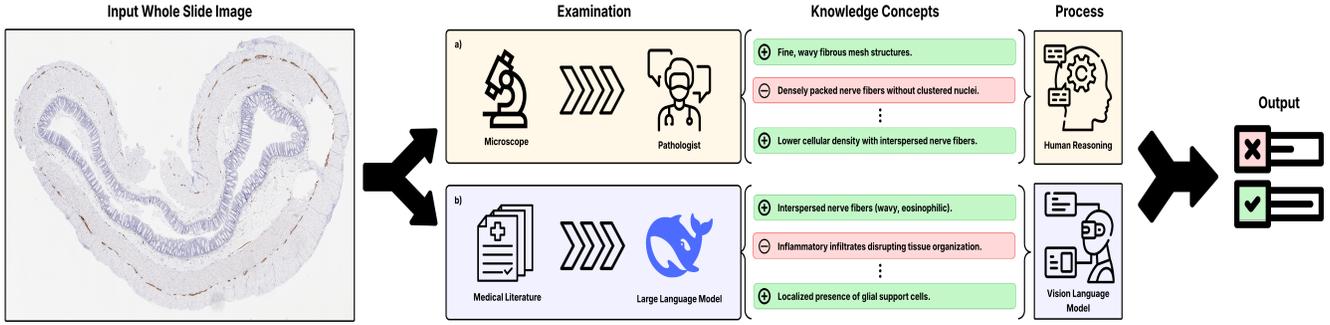

Figure 3: A) The real clinical processes. B) Utilizing an LLM to induce expert concepts related to diagnosis from medical literature.

to distinguish between plexus and non-plexus regions within the muscularis propria layer.

### 2.3 CNN Models

We fine-tuned a pre-trained CNN model called VGG-19 as a baseline. It has five sequential blocks made up of 3×3 conv filters followed by 2×2 maxpool layers with three fully connected layers at the end [19]. VGG-19 showed state-of-the-art performance on ImageNet large scale image classification tasks [20]. Therefore, VGG-19 serves as a strong baseline in computer vision classification tasks. In addition, we evaluated two variants of the Residual Network (ResNet) pre-trained models: ResNet-18 and ResNet-50. All three models are considered vision-based classifiers that are not domain-specific and do not emulate human reasoning during diagnosis. In contrast to VLMs, which integrate multi-modal information with expert-derived concepts.

### 2.4 CLIP Models

The framework (Fig. 4), inspired by [12], combines domain-specific expertise with histopathological image representations using VLMs, particularly those based on the CLIP architecture. CLIP-based multi-modal embeddings and expert-derived textual concepts are used to simulate human reasoning (Fig. 3).

This framework relies on LLMs to extract and refine expert-level textual descriptions from curated medical literature. Instead of acting as knowledge databases, LLMs are used as reasoning engines to generate disease-specific concepts for plexus identification [12, 13]. Using the LLM as a reasoning machine reduces limitations like hallucination (generating unsupported claims) [14] by relying on structured expert knowledge (Fig. 3).

The expert concepts and class prompts for Hirschsprung's Disease were gathered from reputable medical literature (e.g., pathology textbooks, peer-reviewed journals) [5, 6, 17, 18]. The concepts and prompts were extracted using four LLMs: GPT-4o, GPT-o3-mini-high, DeepSeek-R1, and Grok-3. DeepSeek-R1 produced the most clinically coherent and contextually accurate outputs after comparative testing of multiple LLMs, as validated by A. Chan during prompt induction.

After generating expert concepts, the framework embeds image patches and concept prompts in a shared latent space using Quilt-Net. A CLIP-based pathology model pre-trained on over 1 million histopathology image-text pairs is used [15]. Visual embeddings are created from extracted tiles (224×224 pixels), and textual phrases are encoded using a text encoder. The similarity between visual and expert-derived features is then quantified using cosine similarity, which measures the angular proximity between their shared latent space embeddings [12, 16].

The framework uses a two-stage hierarchical aggregation mechanism (Fig. 4): Concept-Level Aggregation and Bag-Level Aggregation. The first stage involves aggregating instance-level features (patch embeddings) into concept-specific bag-level representations using a similarity-based attention mechanism informed by both data-driven and expert-derived concepts [12]. The concept-specific features are combined into a comprehensive slide-level representation in the second stage [12]. Calculating the correlation between bag-level expert class prompts and concept-level embeddings captures higher-level diagnostic context throughout the slide.

Slide adapters (Fig. 4) [12] are used to reduce domain shifts between pre-trained encoders and downstream histopathology classification tasks. These learnable bottleneck layers are placed between the slide representation and the classification head [18]. Slide adapters have two main uses. They adapt the feature distribution to match the target domain and then enable residual-style feature blending to add task-specific signals to the embedding. The lightweight adjustment mechanism enhances generalization without full fine-tuning of the pre-trained vision-language backbone [12].

### 2.5 Training Setup

For training both the CNN and CLIP-based models, we used a consistent training setup for a fair comparison with 5-fold cross-validation, splitting the 30 WSIs. Each fold had four groups (24 WSIs) for training, and a single group that was split in half for validation (3 WSIs), and testing (3 WSIs). These WSIs were tiled into overlapping smaller sub-images of size 224×224 pixels. From each WSI in the training set, 500 tiles were randomly sampled (250 with plexus, 250 with no plexus pixels), resulting in a total of 15,000 tiles.

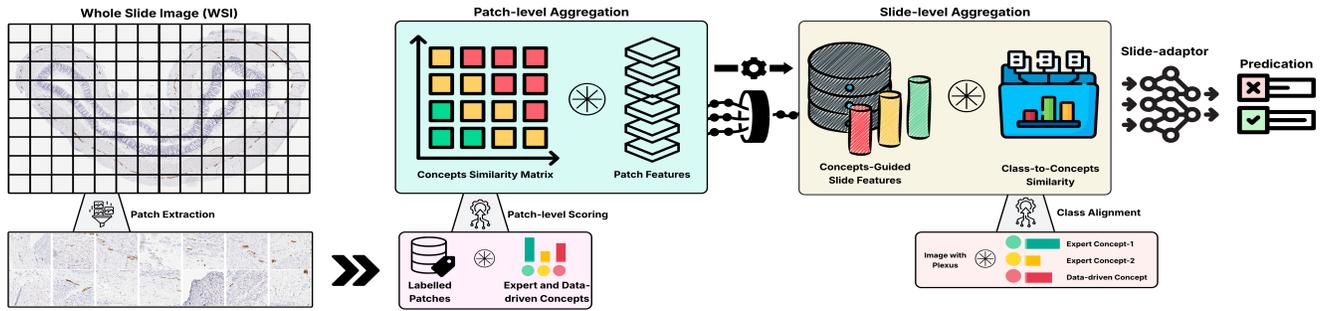

**Figure 4: Framework decomposes a specific complex WSI analysis task into multiple subtasks of scoring patch-level concepts, inspired from [12].**

Data augmentation techniques were applied during training, including random rotations, horizontal and vertical flips, dropout, and scaling for both models. Both were fine-tuned with their hyperparameters using grid search.

We initialized the VGG-19 model with ImageNet-1k pre-trained weights and AdamW optimizer with a weight decay of 1e-4 and a learning rate of 5e-4 were used during training. A cosine learning rate scheduler was applied with a warm-up period of five epochs, after which the learning rate was gradually reduced over 20 epochs with a batch size of 64.

The CLIP-based model used a two-stage hierarchical aggregation approach to leverage visual data and DeepSeek-R1-derived expert concepts. QuiltNet vision-language model uses ViT-B/32 image encoder [11] and GPT/77 text encoder. The training incorporated instance-level and bag-level expert concepts, represented through 16 learnable tokens, alongside one instance-level concept for each target classification class. The model was trained using an AdamW optimizer with a learning rate of 1e-4. Additional hyperparameters were tuned, such as 8 data-driven concepts, an orthogonal ratio of 2, and 16 learnable context tokens.

## 3 Results

Table 1 provides a summary of the performance metrics for the CNN models and the proposed CLIP-based (QuilNet) model. With an accuracy of 83.93%, precision of 86.61%, specificity of 87.60%, F1 Micro and Macro scores of approximately 84%, and an AUC of 91.76%, the QuiltNet model outperformed the others in the majority of metrics. Conversely, the CNN-based models have performed worse on every metric except recall; both ResNet-50 and VGG-19 had higher recall (sensitivity) scores with 86.67% and 81.20%, respectively, at the cost of much lower precision and specificity.

Confusion matrices provided further insights into the model performances (Fig. 5a and 5b). QuiltNet correctly classified 3,009 true positives and 3,285 true negatives, with 465 false positives and 741 false negatives. In contrast, VGG-19 achieved 3,045 true positives and 2,910 true negatives, but at the cost of a higher number of false positives (840).

Adding expert-written text descriptors to the QuiltNet model makes it easier to understand and enhances the automated analysis, making it more similar to the way pathologists think when they examine a case. Examples of these text descriptors, shown in Table 2.

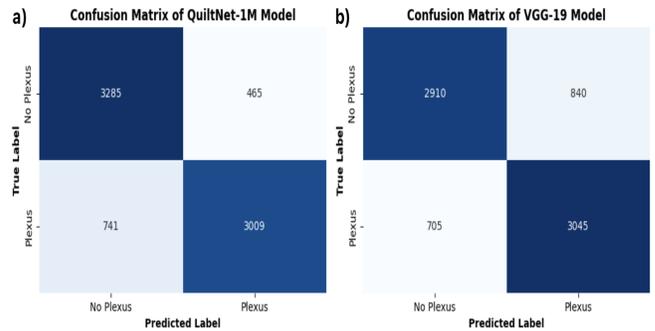

**Figure 5: Confusion matrix of a) QuiltNet model (proposed), b) VGG-19 model (baseline).**

This multimodal approach is similar to how pathologists diagnose patients in real life, where understanding the context and shape of a problem greatly affects how accurate the diagnosis is.

## 4 Discussion

This study proposes a new method for diagnosing HD in colon histological images by identifying plexus areas. Expert-derived textual concepts can be added to a CLIP-based VLM to mimic pathologists' sophisticated reasoning, which utilizes visual signals and contextual information to facilitate accurate diagnosis.

Domain-specific information in the model fixes a major flaw in visual-only automated diagnostic systems. This multimodal approach improves model interpretability and clinical decision-making. Thus, the proposed method may help pathologists identify patients more accurately and reliably, improving patient outcomes.

This successful HD diagnosis suggests that the method could be used for other specialized medical imaging tasks. This study shows that image analysis can use textual terms, enabling computational pathology and better diagnostics.

## 5 Conclusion

In this study, a multi-modal framework is explored to classify Hirschsprung's Disease with expert-derived knowledge concepts to emulate the clinical diagnosis process using a CLIP-based VLM.

Table 1: Comparison of model performance metrics between VGG-19 (baseline) and QuiltNet (proposed).

| Models | Accuracy (%) | Precision (%) | Recall (%) | Specificity (%) | F1 Micro (%) | F1 Macro (%) | AUC (%) |
|---|---|---|---|---|---|---|---|
| VGG-19 (Baseline) | 79.40 | 78.38 | 81.20 | 77.60 | 79.40 | 79.02 | 89.13 |
| ResNet-18 | 58.91 | 57.60 | 67.52 | 50.29 | 58.91 | 55.74 | 64.12 |
| ResNet-50 | 56.95 | 54.36 | **86.67** | 27.23 | 56.94 | 51.53 | 64.67 |
| **QuiltNet (Proposed)** | **83.93** | **86.61** | 80.24 | **87.60** | **83.93** | **83.86** | **91.76** |

Table 2: Examples of expert-derived text descriptors generated with DeepSeek-R1.

| Label | Text Descriptor |
|---|---|
| Plexus (+) | + Clustered large nuclei with prominent nucleoli: Fine, wavy fibrous mesh structures. |
| | + Clustered large neuronal cell bodies with prominent nucleoli: Reticular stromal fibre networks. |
| No Plexus (-) | - Absence of clustered ganglion cells: Uniform smooth muscle layers lacking neural structures. |
| | - Enlarged, densely packed nerve fibres without clustered nuclei. |

The proposed model demonstrated higher performance across most metrics in comparison to the CNN models. Limitations such as restricted prompt diversity and dataset scale currently constrain the full potential of the proposed method, which will be addressed in future work. To help with reproducibility and more research in this area, we have made the full implementation of this work, including code, model configurations, and concept prompts, available to the public on our [GitHub Repository](https://github.com/Yusufii9/KDVLM/tree/main)
(https://github.com/Yusufii9/KDVLM/tree/main).